\documentclass[conference]{IEEEtran}
\IEEEoverridecommandlockouts
\usepackage{cite}
\usepackage{amsmath,amssymb,amsfonts}
\usepackage{algorithmic}
\usepackage{graphicx}
\usepackage{textcomp}
\usepackage{xcolor}
\def\BibTeX{{\rm B\kern-.05em{\sc i\kern-.025em b}\kern-.08em
    T\kern-.1667em\lower.7ex\hbox{E}\kern-.125emX}}
\usepackage{hyperref}
\begin{document}

% --- 1. FILL IN YOUR TITLE HERE ---
\title{Unlocking the Black Box: A Five-Dimensional Framework for Evaluating Explainable AI in Credit Risk\\
}

% --- 2. FILL IN YOUR AUTHOR INFO HERE ---
% --- Delete any unused \and \IEEEauthorblockN blocks ---
\author{\IEEEauthorblockN{1\textsuperscript{st} Rongbin Ye}
\IEEEauthorblockA{\textit{Independent Researcher} \\
Chicago, USA \\
rongbin.ye.94@gmail.com}
\and
\IEEEauthorblockN{2\textsuperscript{nd} Jiaqi Chen}
\IEEEauthorblockA{\textit{Independent Researcher} \\
Chicago, USA \\
ronanchen0901@gmail.com}
}

\maketitle

\begin{abstract}
The financial industry faces a significant challenge modeling and risk portfolios: balancing the predictability of advanced machine learning (ML) models, neural network models, and explainability required by regulatory entities (such as Office of the Comptroller of the Currency (OCC), Consumer Financial Protection Bureau (CFPB). This paper intends to fill the gap in the application between these "black-box" models and explainability frameworks, such as LIME and SHAP. Authors elaborate on the application of these frameworks on different models and demonstrates the more complex models with better prediction powers could be applied and reach the same level of the explainability, using SHAP and LIME. Beyond the comparison and discussion of performances, this paper proposes a novel five-dimensional framework evaluating Inherent Interpretability, Global Explanations, Local Explanations, Consistency, and Complexity—to offer a nuanced method for assessing and comparing model explainability beyond simple accuracy metrics. This research demonstrates the feasibility of employing sophisticated, high-performing ML models in regulated financial environments by utilizing modern explainability techniques and provides a structured approach to evaluate the crucial trade-offs between model performance and interpretability.
\end{abstract}


\section{Introduction}
Credit risk assessment remains a fundamental challenge across the financial and credit industry. This challenge is not confined to traditional banks (e.g., Discover, Bank of America, American Express) \cite{akhter2021comparison}, but is also a critical concern for FinTech leaders such as PayPal and Block \cite{gambacorta2019fintech}. It even extends to "BigTech" platforms like Uber and Airbnb, whose business models increasingly intersect with financial services \cite{frost2021bigtech}. For this diverse set of organizations, credit risk is a foundational pillar challenging profitability and risk-control capabilities. Therefore, the ability to accurately estimate potential exposure and measure its impact on financial resiliency is key to navigating macroeconomic volatility and shifts in consumer confidence.

To address credit assessment, different advanced algorithms and models were investigated and deployed. From the machine learning perspective, the models with higher complexity and designs are prone to capture more complicated relationships, thus capturing subtle signals for potential losses. 

However, this pursuit of predictive power is constrained by strict legal and compliance requirements. Regulators, including the Office of the Comptroller of the Currency (OCC) and the Consumer Financial Protection Bureau (CFPB), mandate that models be accountable, fair, and transparent \cite{occ2022model}. This mandate presents two primary challenges. First, regulations like the Fair Credit Reporting Act (FCRA) prohibit the use of protected attributes (e.g., race, gender, age) to prevent systemic bias. Second, the "black-box" nature of complex models conflicts with regulatory demands for interpretability, which is necessary for providing legally required adverse action notices. This conflict has historically restricted the adoption of high-performing models, such as neural networks, in favor of simpler, more transparent alternatives. The recent development of robust explainable AI (XAI) frameworks, however, presents an opportunity to resolve this tension. It may now be feasible to deploy advanced models and provide the necessary explanations to satisfy regulatory scrutiny.
This research paper aims to address these challenges by applying and discussing explainable machine learning models for credit risk assessment using the Prosper Marketplace dataset. Prosper Marketplace, as a pioneer in peer-to-peer lending, provides a rich data set that contains borrower characteristics, loan terms, and performance outcomes that is ideal for this analysis. Leveraging modern explainability frameworks of SHAP (SHapley Additive exPlanations) and LIME(Local Interpretable Model-agnostic Explanations), a set of transparent credit risk models of different complexities were developed and tested. 

The rise of AI-driven lending decisions has intensified regulatory scrutiny and calls for "explainable AI" in financial services. This project addresses this timely need by demonstrating how complex machine learning pipelines can be made interpretable without sacrificing performance, potentially opening pathways for financial institutions to adopt more sophisticated models while remaining compliant with regulatory requirements. 

\section{Key Challenges and Research Scope}
The adoption of complex, high-performing "black-box" models in the financial industry is hindered by a fundamental trade-off between predictive power and regulatory compliance. This tension manifests in three primary domains:

\begin{itemize}
    \item Regulatory Constraints: Financial institutions operate under strict oversight from bodies like the OCC and CFPB \cite{occ2022model}. These regulations mandate model fairness and transparency, prohibiting the use of protected attributes to prevent discrimination and requiring clear explanations for credit decisions. This environment implicitly favors traditional, inherently interpretable models over more complex, opaque alternatives.

    \item Business and Performance Impact: This regulatory friction creates a difficult choice. While advanced models (e.g., neural networks) can capture complex, non-linear patterns to improve predictive accuracy, institutions are often hesitant to deploy them. This results in a direct sacrifice of potential performance (and thus, profitability and risk-control precision) to ensure regulatory adherence.

    \item Stakeholder Trust and Communication: The use of opaque models erodes trust. Regulators and consumers are increasingly demanding transparency in automated decisions. This places a significant communication burden on technical teams, who must be able to justify complex model behavior to non-technical stakeholders, including auditors, compliance officers, and customers.
\end{itemize}

This paper focuses on developing and evaluating explainable machine learning models to address this specific trade-off, using the Prosper Marketplace dataset as an exemplary case. Our objective is to demonstrate the application of XAI frameworks (LIME mainly) to bridge the gap between high-performance models and regulatory explainability requirements. Accordingly, the deployment of production level risk scoring strategies and pipeline, integrated analysis with products or platform, and real-time scoring system is not part of this paper.

The main goal of this paper is gain a comprehensive understanding of the applicability of different explainability frameworks on different machine learning models. Then, based on the comparison in this case studies, Authors intend to discuss the building of a scalable framework of explainability that could be applied when communicating with different stakeholders for different purposes. 

\section{Literature Review}
This review provides a foundation for our research by surveying literature in three key areas: (1) the technical methodologies of prominent explainable AI (XAI) frameworks; (2) prior applications of these frameworks within the financial industry; and (3) the regulatory landscape that necessitates model explainability.

\subsection{Methodologies of Explainable Frameworks}
The pursuit of model explainability is central to the practical application of machine learning. While foundational texts by James et al. \cite{james2013introduction} and Xu \cite{xu2023machine} establish the modern ML pipeline, the value derived from these systems is contingent on understanding their outputs. Consequently, a rich field of XAI has emerged.

Seminal work by Lundberg and Lee \cite{lundberg2017unified} introduced SHAP (SHapley Additive exPlanations), a game-theoretic approach that unified several existing methods. Building on the early application of Shapley values to regression analysis by Lipovetsky and Conklin \cite{lipovetsky2001analysis}, SHAP provides a robust method for assigning fair, additive feature importance values for any model's prediction. It has become a standard for both global and local model explanations.

For local, model-agnostic explanations, Ribeiro et al. \cite{ribeiro2016should} developed LIME. LIME explains individual predictions by learning a simpler, interpretable surrogate model (e.g., a linear model) in the local vicinity of the prediction, answering the question, "Why did the model make this specific decision for this instance?"

Other key frameworks address specific model types or explanation goals. For deep neural networks, Layer-Wise Relevance Propagation (LRP) by Bach et al. \cite{bach2015pixel} provides granular, pixel-wise explanations by propagating relevance scores backward through the network layers. Frameworks like Quantitative Input Influence (QII) \cite{datta2016algorithmic} and MMD-critic \cite{kim2016examples} offer alternative approaches to global model understanding through causal attribution and prototype-based explanations, respectively. Finally, theoretical work by Miller \cite{miller2019explanation} integrates insights from the social sciences, arguing that for explanations to be effective, they must be contrastive, selective, and socially situated---a principle guiding human-centric XAI design.

\subsection{Applications of Explainability in Finance}
The application of XAI in finance is accelerating as firms seek to deploy more complex models in high-stakes environments. Bracke et al. \cite{bracke2019machine} applied Shapley values to a mortgage default model, validating the method by showing that its findings on key risk drivers (e.g., loan-to-value ratios) aligned with established economic theory. While not using a formal XAI method, the work of Giesecke et al. \cite{giesecke2020deep} on deep learning for mortgage risk underscores the critical need for interpretability when dealing with massive, complex datasets. More recently, Kumbhar et al. \cite{kumbhar2024ai} implicitly highlighted the importance of explainability for ensuring fairness in AI-driven credit scoring systems for underserved populations.

\subsection{Regulatory Frameworks and Implications}
The adoption of XAI in finance is not merely a technical choice but a regulatory necessity. The Office of the Comptroller of the Currency (OCC) mandates rigorous model risk management, including documentation, validation, and governance, which directly incentivizes the use of explainable models \cite{occ2022model}. Furthermore, the need for human-centric, justifiable explanations, as explored by Miller \cite{miller2019explanation}, aligns directly with regulatory expectations for transparency and accountability in consumer finance. This paper builds upon this existing work by applying leading XAI frameworks to a modern P2P lending dataset, with the goal of providing a structured evaluation of the trade-off between model performance and explainability.

\section{Methodology}
This research employs a mixed-methods approach to evaluate the trade-off between model performance and explainability in credit risk assessment. The methodology integrates quantitative analysis of ML models with a qualitative evaluation of their explanation outputs, guided by established frameworks and regulatory standards.

\subsection{Analytical Process}
Our methodology follows a four-stage process, guided by the ML pipelines described by James et al. \cite{james2013introduction} and Xu \cite{xu2023machine}, and the qualitative analysis framework from Schreier \cite{schreier2012}.

\begin{enumerate}
    \item Model Development:
    A comprehensive machine learning pipeline was developed to predict loan default. Authors implemented three models of increasing complexity to represent the spectrum from interpretable to "black-box" systems: 
    \begin{itemize}
        \item Logistic Regression (baseline interpretable model)
        \item Random Forest
        \item Neural Network
    \end{itemize}

    \item Explainability Implementation:
    Authors applied post-hoc, model-agnostic explainability frameworks to the trained models. This research focuses on two of the most prominent methods: SHAP (SHapley Additive exPlanations) \cite{lundberg2017unified} and LIME (Local Interpretable Model-agnostic Explanations) \cite{ribeiro2016should}.

    \item Quantitative \& Qualitative Evaluation:
    The models were evaluated using a dual-faceted approach. 
    \begin{itemize}
        \item Performance: Authors assessed predictive power using standard classification metrics, including Area Under the Curve (AUC), precision, and recall.
        \item Interpretability: Authors conducted a qualitative analysis of the explanations generated by SHAP and LIME, evaluating their utility, consistency, and alignment with regulatory expectations for transparency.
    \end{itemize}

    \item Framework Synthesis:
    Based on the trade-offs identified, this paper proposes a novel, multi-dimensional framework for evaluating and comparing model explainability. This framework is designed to provide a structured rubric for assessing interpretability beyond simple accuracy metrics, thereby helping institutions justify the adoption of more complex models.
\end{enumerate} 

\subsection{Data}
The primary dataset for this research is the Prosper Loan Data, sourced from Kaggle \cite{okam2022prosper}. As a prominent peer-to-peer (P2P) lending platform, this dataset provides a comprehensive, real-world collection of anonymized borrower characteristics, loan terms, and repayment outcomes. The data quality and structure are highly representative of industry-level loan portfolios, making it an ideal testbed for credit risk modeling. All models were trained and tested using this dataset.

To provide critical economic context and enhance model performance, Authors supplement the loan-level data with macroeconomic time-series variables. These variables, including Gross Domestic Product (GDP), the civilian unemployment rate, and the Housing Price Index (HPI), were sourced from the Federal Reserve Bank of St. Louis (FRED) \cite{fred_data}. The inclusion of these external factors is motivated by established research demonstrating the strong correlation between macroeconomic conditions and credit default rates \cite{giesecke2020deep}. All features were assessed for multicollinearity during preprocessing and pruned as necessary.

\subsection{Dataset and Feature Engineering}
The original Prosper dataset contains approximately 113,000 records and 80 features, capturing borrower attributes, loan characteristics, and internal platform metrics \cite{okam2022prosper}.

\subsubsection{Target Variable}
The predictive task is a binary classification of credit default. The target variable, a boolean \texttt{Default} indicator, was engineered from the original \texttt{LoanStatus} feature. Loans with statuses such as 'Chargedoff' or 'Defaulted' were classified as default (\texttt{1}), while those 'Completed' or 'Current' were classified as non-default (\texttt{0}).

\subsubsection{Feature Selection}
A curated set of continuous and categorical variables was selected to evaluate default likelihood. These features are grouped into three primary categories:
\begin{itemize}
    \item Borrower and Loan Attributes: Key financial metrics at origination (e.g., \texttt{StatedMonthlyIncome}, \texttt{DebtToIncomeRatio}) and loan terms (e.g., \texttt{LoanOriginalAmount}, \texttt{InterestRate}).
    \item Credit and Behavioral Metrics: This includes "off-us" data, such as the borrower's external credit score, and "on-us" platform-specific data, such as the internal \texttt{ProsperScore}.
    \item Macroeconomic Variables: External factors (e.g., GDP, unemployment rate) from the Federal Reserve Bank of St. Louis (FRED) \cite{fred_data} were merged to capture systemic economic risk.
\end{itemize}

\section{Findings \& Analysis}
According to the analysis plan, the following section will discuss the EDA results, Modeling Performance, and the applicable of the framework. 

\subsection{Data Presentation - Categorical}
Within the data, there are a few categorical data that this paper will leverage in the model progress. These categorical variables will use one-hot encoding and generating the columns.

\begin{figure}[htbp]
\centerline{\includegraphics{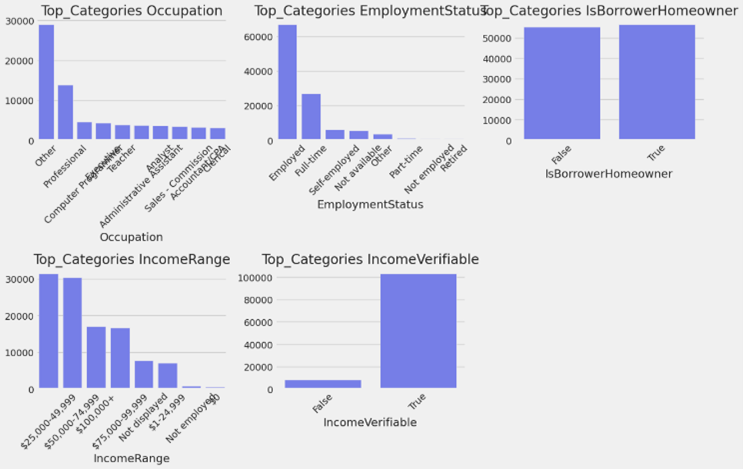}}
\caption{Distribution of Key Categorical Features.}
\label{fig:Categorical}
\end{figure}

After examination, there is no obvious imbalance treatment or extreme outliers require systematic treatment accordingly as shown in Fig. \ref{fig:Categorical}. The one exception, IncomeVerfiable, shows slightly imbalanced pattern. Nevertheless, since most of the data is skewed to verifiable, there should be no significant negative impact that could change the dynamic of the data or research.  

The date credit pulled is in the date time perspective and the same for the first recorded credit line. In this case, Authors conducted further data manipulation to clean up these two columns. 

\subsection{Data Presentation - Continuous}
\begin{figure}[htbp]
\centering % <-- Use \centering instead of \centerline
\includegraphics[width=0.4\textwidth]{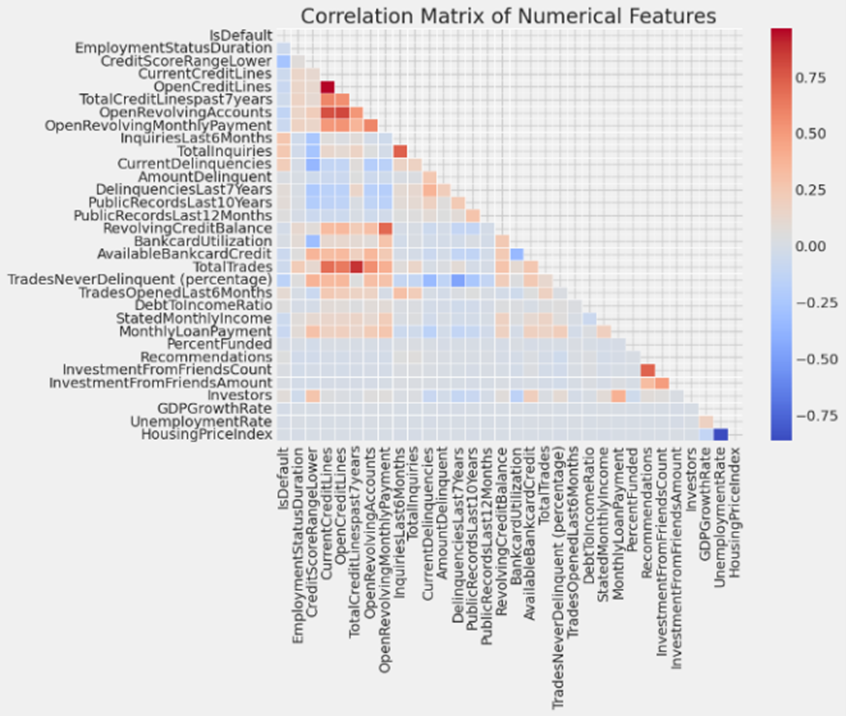} % <-- CONTROLS THE SIZE
\caption{Treatment of Continuous Data (Before)}
\label{fig:before}
\end{figure}

\begin{figure}[htbp]
\centering
\includegraphics[width=0.4\textwidth]{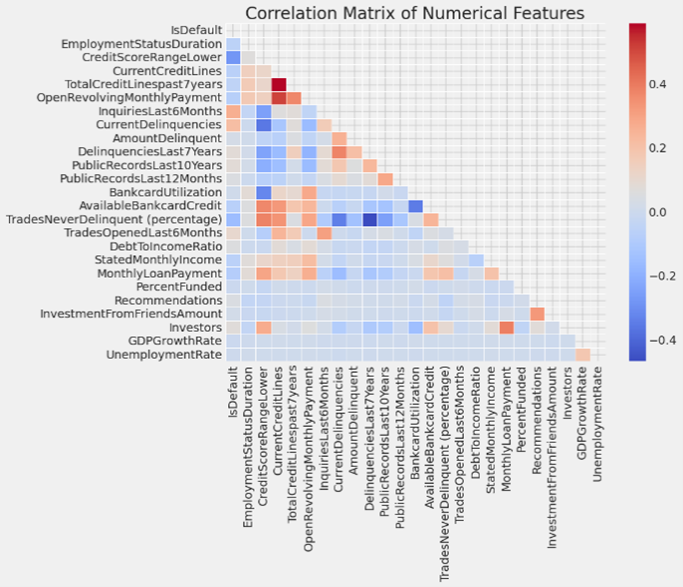} 
\caption{Treatment of Continuous Data (After)}
\label{fig:after}
\end{figure}

The continuous data was first analyzed for multi-collinearity. To ensure model stability and reduce redundancy, features with high inter-correlation were pruned. All remaining continuous variables were then standardized using z-score normalization to ensure they shared a common scale for model training. The effect of this transformation on the feature distributions is visualized in Fig. \ref{fig:before} (before) and Fig. \ref{fig:after} (after).

Following this preprocessing progress, the three models were trained. As hypothesized, a clear performance hierarchy emerged, with predictive accuracy scaling directly with model complexity.

\subsection{Model Implementation}
Three models were trained on the final dataset, representing a spectrum of complexity from traditional statistical methods to deep learning.

\begin{itemize}
    \item Logistic Regression (LR): A standard Logistic Regression model was implemented as a highly interpretable baseline, which is a classical approach for binary classification and predicting probabilities.

    \item Random Forest (RF): An ensemble classifier was built using a Random Forest. The model was optimized via grid search, resulting in a final configuration of 100 trees (\texttt{n\_estimators=100}). It employs \texttt{max\_features='sqrt'} for split decisions and \texttt{class\_weight='balanced'} to counteract dataset imbalance.
    
    \item Neural Network (NN): A sequential, feedforward Neural Network was constructed. The architecture consists of three hidden layers (64, 32, and 16 neurons) with 'relu' activation, followed by a final 'sigmoid' output layer for binary classification. To prevent overfitting, two dropout layers (rates of 0.3 and 0.2) were incorporated, and an early stopping callback was set to monitor validation loss with a patience of 5 epochs. The model was compiled using the 'adam' optimizer and 'binary\_crossentropy' loss function.
\end{itemize}

\subsection{Comparative Analysis}
\begin{figure}[htbp]
\centerline{\includegraphics{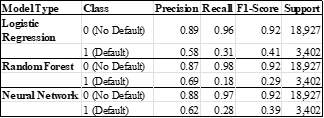}}
\caption{Comparison of Performances}
\label{tab:model_performance}
\end{figure}

As designed, the models represent a clear hierarchy of increasing complexity: $LR < RF \ll NN$. This hierarchy in complexity correlated directly with predictive performance, as detailed in Fig. \ref{tab:model_performance}.

The Neural Network yielded the best performance, as hypothesized. This "lift" was primarily driven by its superior precision and F1-score on the critical default (minority) class. While all three models performed comparably well at identifying non-default cases, the NN's ability to model complex, non-linear relationships translated into a more reliable identification of high-risk loans. This result demonstrates a clear and significant performance gain by the most complex model on the most crucial aspect of the prediction task.

\section{Discussion}

\subsection{Evaluation of Outcome}

\begin{figure}[htbp]
\centerline{\includegraphics{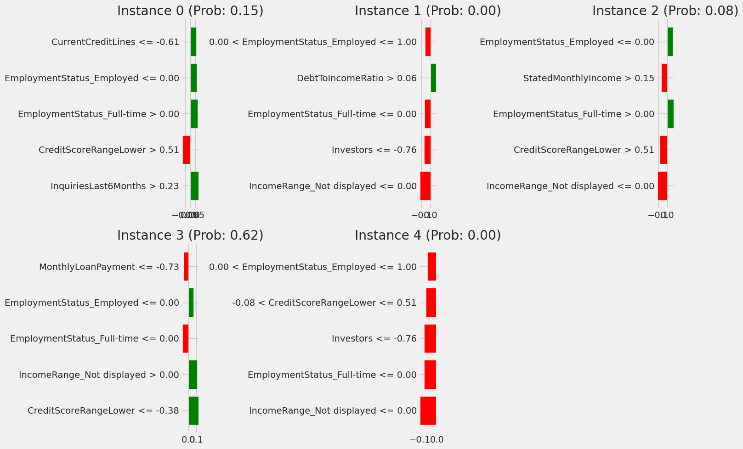}}
\caption{Instance-wise explanation of Neural Network Model}
\label{fig:instance}
\end{figure}

One of the greatest breakthroughs is the explainability over each of the instances, where one could tell what has been launched and what the driving reason for not being selected or selected was in this case. This will be the major contribution that while having a more complicated model, one could still have a model that explains and could be breakdown. Based on the Fig. \ref{fig:instance}., for example, the financial company could provide the legal required letters about the 

\subsection{Recommendation}
After exploration of the major framework, this paper will focus on the aspect of the LIME framework. From the statistical learning perspective, the explanation for the rule-based method is straightforward. Each unit change of the  there is no method to break down the single factor’s impact on the final decision or model results, the probability of default in this case. 

Development of the combination view based on the performance and explainability 
Based on the understanding of the subject of the explainability, Inherent Interpretability, Global Explanations, Local Explanations, Consistency, and Complexity. These five components will consist of a comprehensive approach considering both statistical reference standards and the CFPB rubrics. Instead of understanding the explainability from one perspective, this aggregated collection focused on how well model decisions can be understood from the perspectives of different stakeholders. 

Each of the components will focus on the different aspects of the explainability per se. The inherent interpretability lays the basic transparency of the entire model architecture—whether the mathematical structure itself is understandable for a normal analyst without additional tools or informative teaching. This dimension will address the trade-off between model sophistication and natural understandability. The global explanations measure the possibility to provide overarching insights on feature importance and decision logic, crucial for model documentation and regulatory approval processes as OCC (2023) outlined. The local explanations emphasize the individual instances as previously discussed in the neural network example, which is critical in credit denial justifications and is the key to comply with CFPB requirements. Next, consistency ensures the same explanation will be applied across similar cases, a dimension often overlooked but critical for regulatory trust and operational deployment. Without consistent explanations, stakeholders may question the reliability and potential of violation of the fair credit Act. Finally, Complexity focuses on the nature of the model and result, whether these can be communicated to non-technical stakeholders, especially when some prediction is not working or has some edge cases. 

This balanced framework is designed majorly for credit risk assessment as it balances technical accuracy with practical communication needs with different stakeholders, such as clients, regulators or internal audit department. By weighing these dimensions differently (with higher importance on Local Explanations for credit applications), one can tailor explainability evaluation to the specific requirements. This approach enables organizations to move beyond simplistic accuracy-vs-transparency dichotomies toward a nuanced understanding of explainability that satisfies both performance goals and regulatory mandates. 

\begin{figure}[htbp]
\centerline{\includegraphics{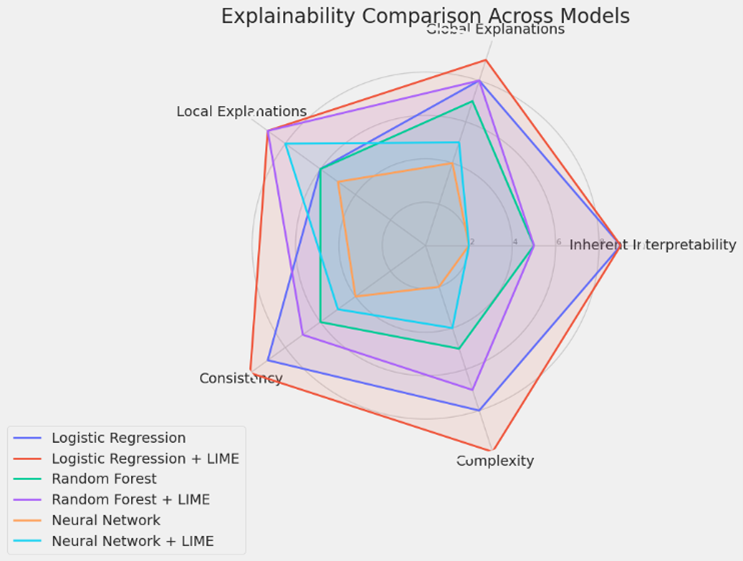}}
\caption{Demonstration of the multiple dimension of the
explainability}
\label{fig:Demention}
\end{figure}

To facilitate how this framework could be applied, Fig. \ref{fig:Demention} is an example that modeler or model risk or regulator can clearly tell the combination score with the explainability on the selected model.

\subsection{Future Improvement and Next Steps}
The findings of this study, while demonstrative, are subject to several limitations that provide clear avenues for future research.

The scope of this project necessitated a focused analysis on the application of the LIME and SHAP frameworks to a single dataset. While our results confirmed that more complex models can achieve superior predictive performance under identical computational constraints, a more rigorous, quantitative comparison of the explainability frameworks themselves (e.g., evaluating explanation stability, fidelity, and computational cost) was not performed.

Future work should expand upon this foundation in several key directions:
\begin{itemize}
    \item Framework Comparison: A more comprehensive, head-to-head analysis of LIME and SHAP, evaluating the relative strengths and weaknesses of their explanations across all tested model types.
    
    \item Methodological Validation: Applying the proposed five-dimensional explainability framework to a wider array of larger, more diverse financial datasets to test its generalizability and robustness.
    
    \item Advanced Architectures: Extending this explainability analysis to state-of-the-art models. A significant future contribution would be the evaluation of XAI techniques on transformer-based deep learning structures, including Large Language Models (LLMs), to address the explainability of these more advanced "black-box" systems in credit risk assessment.
\end{itemize}

\section{Conclusion}
In summary, this paper utilized the Prosper Marketplace loan dataset to develop and compare three distinct classification models. By applying the LIME explainability framework, the paper demonstrated that it is feasible to deploy complex, high-performance models (such as neural networks) while still providing the detailed, instance-wise explanations required by regulatory and business stakeholders.

Beyond this practical demonstration, the paper's primary contribution is the proposal of a novel five-dimensional framework for evaluating model explainability. This framework which assesses Inherent Interpretability, Global Explanations, Local Explanations, Consistency, and Complexity—provides a structured rubric for a more nuanced comparison of models, moving beyond the simple trade-off between accuracy and interpretability. Authorsposit that this framework can serve as a cornerstone for the justified and responsible adoption of more sophisticated models in regulated financial environments.

Future work should focus on validating this framework across a wider range of XAI methods, including SHAP, and applying it to other complex financial datasets. The continued development and structured evaluation of explainable AI are critical for driving innovation and ensuring transparency in the credit industry.


\begin{thebibliography}{00}

\bibitem{akhter2021comparison}
T. M. Akhter, "A comparative analysis of credit risk management between traditional banks and fintech lenders," \textit{International Journal of Economics, Commerce and Management}, vol. IX, no. 6, pp. 240-254, 2021.

\bibitem{gambacorta2019fintech}
L. Gambacorta, Y. Huang, H. S. Shin, and H. Zhou, "Fintech and credit scoring," Bank for International Settlements (BIS), Working Papers No. 838, 2019. [Online]. Available: https://www.bis.org/publ/work838.pdf

\bibitem{frost2021bigtech}
J. Frost, L. Gambacorta, Y. Huang, H. S. Shin, and P. Zbinden, "BigTech and the changing structure of financial intermediation," \textit{Economic Policy}, vol. 36, no. 108, pp. 761–807, 2021. doi: 10.1093/ep/eiab023

\bibitem{occ2022model}
{Office of the Comptroller of the Currency}, "Model Risk Management (Comptroller's Handbook)," Office of the Comptroller of the Currency, 2022.

% --- Paste these entries into your \begin{thebibliography}{99} block ---

\bibitem{james2013introduction}
G. James, D. Witten, T. Hastie, and R. Tibshirani, \textit{An Introduction to Statistical Learning}. New York, NY, USA: Springer, 2013.

\bibitem{schreier2012} M. Schreier, \textit{Qualitative Content Analysis in Practice}. London: Sage, 2012. [Online]. Available: http://digital.casalini.it/9781446258750

\bibitem{xu2023machine}
A. Xu, \textit{Machine Learning System Design Interview}. ByteByteGo, 2023.

\bibitem{lundberg2017unified}
S. M. Lundberg and S.-I. Lee, "A unified approach to interpreting model predictions," in \textit{Proc. Adv. Neural Inf. Process. Syst. (NIPS)}, 2017, pp. 4765--4774.

\bibitem{lipovetsky2001analysis}
S. Lipovetsky and M. Conklin, "Analysis of regression in game theory approach," \textit{Applied Stochastic Models in Business and Industry}, vol. 17, no. 4, pp. 319–330, 2001.

\bibitem{ribeiro2016should}
M. T. Ribeiro, S. Singh, and C. Guestrin, "``Why should I trust you?'': Explaining the predictions of any classifier," in \textit{Proc. 22nd ACM SIGKDD Int. Conf. Knowl. Discovery Data Mining}, 2016, pp. 1135--1144.

\bibitem{bach2015pixel}
S. Bach, A. Binder, G. Montavon, F. Klauschen, K.-R. Müller, and W. Samek, "On pixel-wise explanations for non-linear classifier decisions by layer-wise relevance propagation," \textit{PLoS ONE}, vol. 10, no. 7, p. e0130140, 2015.

\bibitem{datta2016algorithmic}
A. Datta, S. Sen, and Y. Zick, "Algorithmic transparency via quantitative input influence: Theory and experiments with learning systems," in \textit{Proc. IEEE Symp. Security and Privacy (SP)}, 2016, pp. 598--617.

\bibitem{kim2016examples}
B. Kim, R. Khanna, and O. O. Koyejo, "Examples are not enough, learn to criticize! criticism for interpretability," in \textit{Proc. Adv. Neural Inf. Process. Syst. (NIPS)}, 2016, pp. 2280--2288.

\bibitem{miller2019explanation}
T. Miller, "Explanation in artificial intelligence: Insights from the social sciences," \textit{Artificial Intelligence}, vol. 267, pp. 1--38, 2019.

\bibitem{bracke2019machine}
P. Bracke, A. Datta, C. Jung, and S. Sen, "Machine learning explainability in finance: An application to default risk analysis," Bank of England Staff Working Paper No. 816, 2019.

\bibitem{okam2022prosper}
H. Okam, "Prosper loan data," [Data set], Kaggle, 2022. [Online]. Available: \url{https://www.kaggle.com/datasets/henryokam/prosper-loan-data}

\bibitem{fred_data}
{Federal Reserve Bank of St. Louis}, "Federal Reserve Economic Data (FRED)," [Data set], 2024. [Online]. Available: \url{https://fred.stlouisfed.org}

\bibitem{giesecke2020deep}
K. Giesecke, J. Sirignano, and A. Sadhwani, "Deep learning for mortgage risk," \textit{Journal of Financial Econometrics}, vol. 19, no. 2, pp. 313–368, 2020.

\bibitem{kumbhar2024ai}
T. Kumbhar, D. Agrawal, L. Saldanha, and D. Koshti, "AI-driven credit scoring and credit line solution for the unreserved and self-employed," in \textit{Proc. 2nd Int. Conf. Inventive Comput. Inf. (ICICI)}, 2024, pp. 178--184.


\bibitem{ref1} A. Adadi and M. Berrada, ``Peeking inside the black-box: A survey on explainable artificial intelligence (XAI),'' \textit{IEEE Access}, vol. 6, pp. 52138--52160, 2018. [Online]. Available: https://doi.org/10.1109/ACCESS.2018.2870052

\bibitem{ref2} R. Adams, V. M. Bord, and B. Katcher, ``Credit card profitability,'' \textit{FEDS Notes}, Board of Governors of the Federal Reserve System, 2022.

\bibitem{ref8} L. M. Demajo, V. Vella, and A. Dingli, ``Explainable AI for interpretable credit scoring,'' \textit{arXiv preprint arXiv:2012.03749}, 2020.

\bibitem{ref9} Drwhy.ai, ``LIME,'' 2025. [Online]. Available: https://ema.drwhy.ai/LIME.html

\bibitem{ref12} B. Kim, R. Khanna, and O. O. Koyejo, ``Examples are not enough: Learn to criticize, criticism for interpretability,'' in \textit{Proceedings of the 30th International Conference on Neural Information Processing Systems}, 2016, pp. 2280--2288.

\bibitem{ref14} LIME, ``GitHub repository,'' 2025. [Online]. Available: https://github.com/marcotcr/lime?tab=readme-ov-file

\bibitem{ref17} McKinsey \& Company, ``Embracing generative AI in credit risk,'' n.d. [Online]. Available: https://www.mckinsey.com/capabilities/risk-and-resilience/our-insights/embracing-generative-ai-in-credit-risk

\bibitem{ref19} B. H. Misheva, J. Osterrieder, A. Hirsa, O. Kulkarni, and S. F. Lin, ``Explainable AI in credit risk management,'' \textit{arXiv preprint arXiv:2103.00949}, 2021.


\bibitem{ref23} Office of the Comptroller of the Currency, ``Credit card lending,'' n.d. [Online]. Available: https://www.occ.treas.gov/publications-and-resources/publications/comptrollers-handbook/files/credit-card-lending/pub-ch-credit-card.pdf

\bibitem{ref25} PYMNTS, ``The black box: When AI calls shots we can't explain,'' 2024. [Online]. Available: https://www.pymnts.com/artificial-intelligence-2/2024/the-black-box-when-ai-calls-shots-we-cant-explain/

\bibitem{ref26} M. T. Ribeiro, S. Singh, and C. Guestrin, ``Model-agnostic interpretability of machine learning,'' \textit{arXiv preprint arXiv:1606.05386}, 2016.


\end{thebibliography}
\end{document}